# OVERHEAD MNIST: A BENCHMARK SATELLITE DATASET

David A. Noever and Samantha E. Miller Noever

PeopleTec, Inc., Huntsville, Alabama, USA
david.noever@peopletec.com

## ABSTRACT

The research presents an overhead view of 10 important objects and follows the general formatting requirements of the most popular machine learning task: digit recognition with MNIST. This dataset offers a public benchmark extracted from over a million human-labelled and curated examples. The work outlines the key multi-class object identification task while matching with prior work in handwriting, cancer detection and retail datasets. A prototype deep learning approach with transfer learning and convolutional neural networks (MobileNetV2) correctly identifies the ten overhead classes with average accuracy of 96.7%. This model exceeds the peak human performance of 93.9%. For upgrading satellite imagery and object recognition, this new dataset benefits diverse endeavors such as disaster relief, land use management and other traditional remote sensing tasks. The work extends satellite benchmarks with new capabilities to identify efficient and compact algorithms that might work on-board small satellites, a practical task for future multi-sensor constellations. The dataset is available on Kaggle and Github.

## 1. INTRODUCTION

The most popular starting test for both new and established machine learning algorithms relies on handwritten digit [1] or letter [2] recognition. If a method does not work with the Modified National Institute of Standards and Technology dataset (MNIST), it most likely will not work on more challenging tasks. As illustrated in Figure 1, the core task corresponds to a multi-class image challenge, one which proves common and useful in other fields [3] outside of algorithms to interpret handwriting. Researchers over the last two decades [4] have spawned more than 48,000 MNIST-related publications, with a quarter of those appearing in 2020. The reverse of this universality stems from the relative ease which modern algorithms have solved the problem (>99% accuracy after a few iterations) [5-6]. Critics of the digit dataset {0-9} note that it contains near-duplicates and lacks diversity in example data such that modifying a single pixel (among 784 pixels in a 28x28 image) can flip some algorithms to misidentify the expected digit [7-9]. Alternative practical extensions of the digit recognition task now include alphabetic handwriting in multiple languages (e.g. English [2], Chinese [10], Russian [11], Kannada [12], American Sign Language [13]) and related everyday object recognition, the most popular of which includes 10 categories of skin cancer [14] and clothing [15] in thumbnail grayscale images (HAM1000 [14] and Fashion-MNIST [15]).

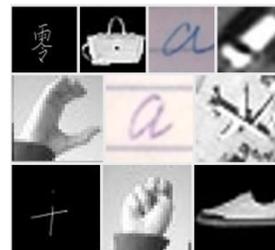

Figure 1. Examples of MNIST families for digits, letters, signing, and objects

The present work provides another challenging object recognition task: labelling objects from overhead satellite imagery (Figure 2). To take advantage of the vast machine learning literature on digit recognition, we mirror the format of the original MNIST closely [1] and thus, like Fashion-MNIST [15], we aim to provide the research community with another drop-in replacement for benchmarking [16]. The grayscale (28x28 pixel) imagery provides a challenging object recognition task [17]. As viewed from above, objects such as planes, ships, and stadiums offer no obvious preferred orientation, so rotating or image shearing may not augment dataset diversity [18]. Overhead-object classifiers also suffer from scale variations that can range more than two orders of magnitude between a small car to a stadium [19-21]. Compared to alternative terrestrial (color) datasets (like CIFAR-10 thumbnail [22]), a classifier of satellite imagery may span different camera resolutions, orientations, and day-night contrast levels in different seasons and shadows. The research offers a novel benchmark [23] for recognizing objects in thumbnail satellite images, a reformatting strategy to connect with the vast MNIST algorithm literature and finally, an initial solution to the classification problem using transfer learning and MobileNetV2 [24]. The original contributions of the present work include: 1) generalizing the standard MNIST format [1,16] and dataset design to handle challenging satellite object detection (called Overhead-MNIST); 2) exploring the unique aspects of overhead object recognition such as diverse object scale lengths and rotational-invariance [17,21]; 3) classifying 10 classes of objects with multiple algorithms, including some efficient enough to run on-board satellites for automated tasking and cueing. Absent the 70,000 overhead thumbnails of the original digit recognition [1], we examine the requirements for solving the problem as a function of sample size.

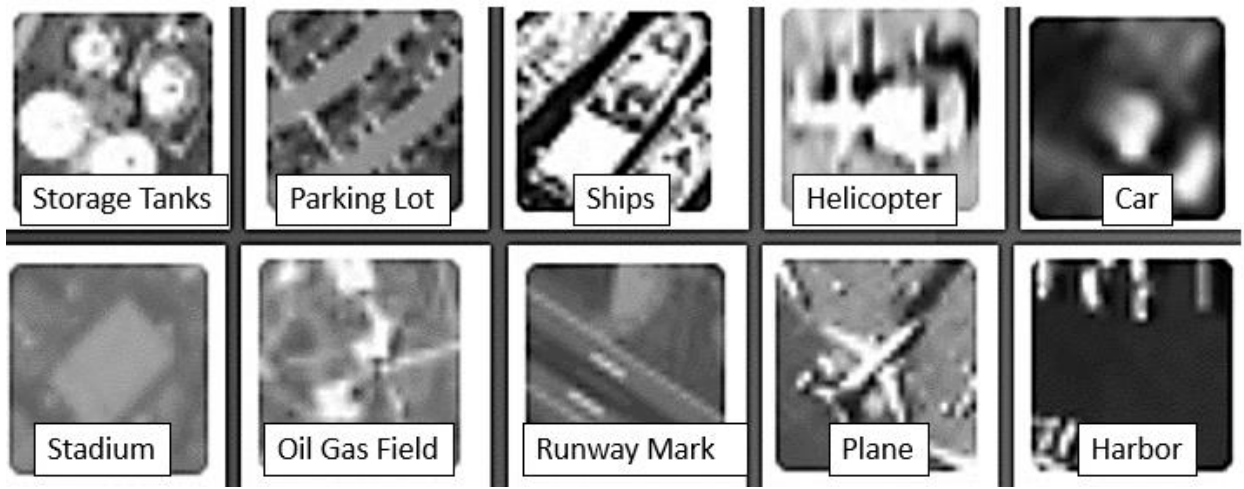

Figure 2. Ten Classes Labelled for Overhead MNIST. *The class selection includes dynamic (car, plane, helicopter, ship) and static (parking lot, runway mark, harbor) targets along with infrastructure-related objects (storage tanks, stadium, oil gas fields)*

## 2. METHODS

We curated the overhead imagery from multiple public sources, xView [17], UC Merced Land Use [25], DOTA [26], and SpaceNet [27]. The cojoined source dataset consists of 102 labelled classes after extraction from any bounding boxes. We resized all the objects to 28x28 pixels in grayscale using ImageMagick [28] command line tools. We approximately balanced (within 20%) the data between the 10 classes with 1000 examples per category. Helicopters are under-represented (800) and harbors are over-represented (1200). For these 10,000 training and testing images (90:10 ratio), the original data was pre-processed as labeled objects in 10 object classes (car, harbor, helicopter, oil gas field, parking lot, plane, runway mark, ship, stadium, and storage tank). While other common classes (like buildings) provide a quick test for urban landscapes, their diversity of scale and overlapping density discouraged their inclusion.

Three different formats are provided for download, including comma-separated values files, JPEG images sorted by class (10 total), and the original MNIST binary format (idx-ubyte) [1]. These three formatting options should cover most all published MNIST solutions with only minor modifications [16]. To mimic the 10-digit recognizer classes. the CSV files for both testing and training have labels in alphabetic order (0-9, where 0=car,1=harbor, etc.). To make a drop-in replacement and establish contact with the existing MNIST datasets, we further process these satellite images into just 10 classes and generate CSV pixel values by converting the gray JPEG to text [28], then parsing out the pixel values in 784 columns for each image as a single labeled row.

The selection of these classes is motivated by their frequent representation among the raw source imagery and diversity among object types of size, shape and background. To examine objects with different natural scale and distinct features, these criteria exclude some common object classes in xView [17], such as buildings, trucks, etc.

The class selection includes a relative balance of dynamic (car, plane, helicopter, ship) and static (parking lot, runway mark, harbor) targets along with infrastructure-related objects (storage tanks, stadium, oil gas fields). This class ontology differs from a standard handwriting digit recognizer (0-9) in diversity [7] and resembles more closely the Fashion-MNIST ontology [15] with shirts, pants, etc. It is worth noting that for these small thumbnails in greyscale at 28x28 pixel, a human analyst might be challenged to distinguish the class differences particularly in their original formats that might include 16 million pixels in the full satellite scan. In the same way that CIFAR-10 dataset [22] provides ambiguous choices at the image size of a thumbnail, this scale also matches what one might encounter in recognizing space objects. For comparison, the 28x28 pixel image of a car in Figure 2 would offer the human analyst approximately 300 pixels to identify a 3x5 square meter object. The base ground sample distance for the original xView [20] and SpaceNet [27] imagery corresponds to WorldView-3 satellite images [21] with a resolution of approximately 0.3-0.7 m per pixel.

Table 1. Results for MobileNetV2 in transfer learning Overhead-MNIST

| Class | Accuracy (Test #) | Class | Accuracy (Test #) |
|---|---|---|---|
| **Plane** | 0.99 (134) | **Oil gas field** | 0.98 (134) |
| **Harbor** | 0.98 (134) | **Parking lot** | 0.96 (134) |
| **Ship** | 0.98 (134) | **Car** | 0.96 (134) |
| **Stadium** | 0.98 (127) | **Helicopter** | 0.93 (99) |
| **Runway** | 0.98 (121) | **Storage tank** | 0.90 (134) |

## 3. RESULTS

As shown in Table 1, a lightweight deep learning model (MobileNetv2) and the Overhead-MNIST dataset can reach 90-100% accuracies on the previously unseen test samples, with storage tanks (90%) the lowest score and planes (99%) the highest score. As shown in Figure 3, the confusion matrix shows correspondence between actual and predicted class assignments in the test set. Current state-of-the-art [6] is 99.84% accuracy for standard digit-recognition on MNIST (using capsule neuronal layer), compared to 96.4% accuracy found here for Overhead-MNIST (using standard convolutional neural networks).

To compare these results across multiple algorithms and feature extraction techniques, we will examine additional benchmarking in a further companion publication. To understand the role of sample number and human comparison, we revisited Fashion-MNIST [15] with its 70,000 images and the same framework for transfer learning (MobileNetV2 [24]).

### 3.1. Comparison with Human Experts

Since humans originally labelled the training and testing data, how could a machine learning algorithm ever exceed human performance? In practice, humans get tired. Not all humans have honed a skill for image analysis or the rapid visual identification of objects, particularly when scoring thousands of tiny thumbnails. Test subjects disagree with the people who labelled the training set. For instance, when compared to a relentlessly attentive machine classifier, a human would have to be 100% right on digit recognition for a thin victory of a few-tenths of a percent.

|  | car | harbor | helicopter | oil_gas_field | parking_lot | plane | runway_mark | ship | stadium | storage_tank | unknown |
|---|---|---|---|---|---|---|---|---|---|---|---|
| car | 129 | 0 | 0 | 0 | 0 | 0 | 1 | 0 | 2 | 2 | 0 |
| harbor | 0 | 131 | 0 | 0 | 1 | 0 | 0 | 2 | 0 | 0 | 0 |
| helicopter | 0 | 0 | 92 | 0 | 0 | 0 | 0 | 0 | 0 | 0 | 0 |
| oil_gas_field | 0 | 0 | 0 | 131 | 1 | 0 | 1 | 0 | 1 | 0 | 0 |
| parking_lot | 0 | 1 | 0 | 2 | 128 | 0 | 0 | 0 | 2 | 1 | 0 |
| plane | 0 | 0 | 2 | 0 | 0 | 132 | 0 | 0 | 0 | 0 | 0 |
| runway_mark | 0 | 0 | 0 | 0 | 0 | 0 | 119 | 0 | 2 | 0 | 0 |
| ship | 0 | 1 | 0 | 0 | 1 | 0 | 1 | 131 | 0 | 0 | 0 |
| stadium | 1 | 0 | 0 | 0 | 0 | 0 | 0 | 0 | 125 | 1 | 0 |
| storage_tank | 0 | 2 | 2 | 0 | 2 | 0 | 0 | 1 | 7 | 120 | 0 |
| unknown | 0 | 0 | 0 | 0 | 0 | 0 | 0 | 0 | 0 | 0 | 20 |

Figure 3. Confusion Matrix using MobileNetV2 and Overhead MNIST. *As a small and fast classifier, the overall average accuracy reaches 96.4% across all object types.*

The more challenging Fashion-MNIST [15] offers a baseline 89.7% accuracy using a statistical technique (Support Vector Classifier, SVC) and greater than 97% using convolutional nets [29]. The comparable human benchmark for 1000 randomly sampled test images was 83.5% accurate among crowd-sourced classifiers (without any fashion expertise) [29]. One might argue the caveat for "no previous experience" limits the human-machine comparison. To supplement this prior work, we worked with one trained volunteer (graduate design degree and fashion experience) and scored 1000 images in a similar test to the published Fashion-MNIST benchmark for human performance. The labeller achieved nearly 100% accuracy in the less ambiguous categories (bag, sneaker) but underperformed the originally labelled data where overlap arises (shirt, coat, T-shirt). Two repetitions of the experimental findings are shown in Figure 4, including a new benchmark human accuracy of 85.12%, which compares will with the 89.7% accuracy of a statistical technique based on feature extraction and image pre-processing. To apply this human labeller as the second opinion on the Overhead-MNIST, Figure 5 summarizes the categorical correctness at 93.86%, with the greatest human ambiguity for parking lots and helicopters. It's worth noting that helicopters and to a lesser extent ships naturally include distortion when transformed from their original rectangular bounding boxes into the square thumbnails for the MNIST format (Figure 2).

| Item | Samples | Correct | Incorrect | Accuracy | Item | Samples | Correct | Incorrect | Accuracy |
|---|---|---|---|---|---|---|---|---|---|
| sandal | 152 | 148 | 4 | 97.37% | sandal | 152 | 149 | 3 | 98.03% |
| pullover | 64 | 63 | 1 | 98.44% | pullover | 64 | 55 | 9 | 85.94% |
| bag | 181 | 181 | 0 | 100.00% | bag | 181 | 181 | 0 | 100.00% |
| shirt | 137 | 79 | 58 | 57.66% | shirt | 137 | 46 | 91 | 33.58% |
| coat | 129 | 66 | 63 | 51.16% | coat | 129 | 67 | 62 | 51.94% |
| tshirt | 226 | 169 | 57 | 74.78% | tshirt | 226 | 217 | 9 | 96.02% |
| sneaker | 108 | 108 | 0 | 100.00% | sneaker | 108 | 107 | 1 | 99.07% |
| trouser | 156 | 155 | 1 | 99.36% | trouser | 156 | 154 | 2 | 98.72% |
| dress | 117 | 110 | 7 | 94.02% | dress | 117 | 105 | 12 | 89.74% |
| total | 1270 | 1079 | 191 | 84.96% | total | 1270 | 1081 | 189 | 85.12% |

Figure 4. Human performance for Fashion-MNIST image classification. *The results highlight variability between ambiguous class identifications that prove hard to decipher from thumbnails in grayscale.*

| Item | Samples | Correct | Incorrect | Accuracy |
|---|---|---|---|---|
| helicopter | 60 | 50 | 10 | 83.33% |
| plane | 108 | 108 | 0 | 100.00% |
| car | 80 | 74 | 6 | 92.50% |
| parking_lot | 88 | 62 | 26 | 70.45% |
| ship | 138 | 134 | 4 | 97.10% |
| storage_tank | 121 | 114 | 7 | 94.21% |
| runway_mark | 114 | 114 | 0 | 100.00% |
| oil_gas_field | 140 | 137 | 3 | 97.86% |
| harbor | 102 | 89 | 13 | 87.25% |
| stadium | 151 | 146 | 5 | 96.69% |
| total | 1042 | 978 | 64 | 93.86% |

Figure 5. Human performance for Overhead-MNIST image classification. *The results highlight variability between ambiguous class identifications that prove hard to decipher from thumbnails in grayscale.*

## 3.2. Dependence on Sample Size

As most practitioners examine overall dataset robustness, accuracies that reach 100% point to over-training, lack of diversity, or insufficient size [7]. Overhead-MNIST offers 9584 labelled thumbnails, or approximately an order of magnitude fewer training instances compared to the original MNIST format and size. The non-digit related MNIST in Chinese [10] and Kannada [12] alternatively currently label 15,000 images total across more classes.

| Item | Samples | Accuracy | Samples | Accuracy | Samples | Accuracy |
|---|---|---|---|---|---|---|
| bag | 147 | 93% | 28 | 96% | 15 | 87% |
| coat | 147 | 77% | 11 | 45% | 11 | 64% |
| dress | 152 | 88% | 21 | 81% | 15 | 73% |
| pullover | 143 | 88% | 21 | 71% | 15 | 73% |
| sandal | 152 | 99% | 23 | 100% | 15 | 100% |
| shirt | 154 | 68% | 17 | 100% | 15 | 93% |
| sneaker | 156 | 97% | 28 | 75% | 15 | 100% |
| trouser | 146 | 97% | 21 | 38% | 15 | 73% |
| tshirt | 146 | 78% | 24 | 100% | 15 | 47% |
| total | 1051 | 87.22% | 149 | 78.44% | 101 | 78.89% |

Figure 6. Effects of sample size on Fashion-MNIST accuracy. *MobileNetv2 and transfer learning shows a drop in accuracy from 150 examples per class but losses plateau at 11 to 30 examples.*

To investigate the differences between these other MNIST datasets (if greater than 70,000 images), we ran additional exploratory experiments on Fashion-MNIST. By shrinking the training data, one can see if the overall accuracy declines once the set reached the 10,000 examples offered by Overhead-

MNIST. Most observers hypothesize that current image classifiers succeed in the range of 100-1000 examples per class, but fewer examples can still work for either transfer learning (10-100 examples) [30] or novel one-shot or few-shot architectures (<20 examples) [31].

Figure 6 summarizes the test results when the training data was varied between 10,000 and 1000 total examples, or approximately 1000-100 per class. The top result for the full 70,000 images (10,000 test) exceed 97%. For this style of diverse thumbnail dataset, one rough figure of merit therefore might consider each order of magnitude increase in training size increases the overall accuracy by around 10% between 1000-70,000 total cases until some dataset-dependent plateaus are reached. We created a larger 76,692 sample of augmented satellite imagery, after applying rotational and brightening transformations. One might hope that to reach superhuman performance [5], an optimal number of images corresponds to a realizable cross-over or goal, particularly given the shortage of more rare or uncollected overhead samples.

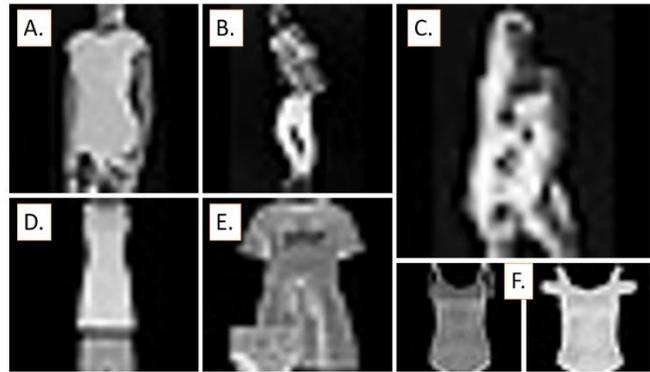

*Figure 7. Very hard or mislabelled thumbnails in Fashion-MNIST. A.T-shirt with person. B. Coat with person C .Dress with person. D. Dress with two parts. E .Dress with bloomers. F. T-shirts.*

### 3.3. Unique Overhead Satellite Features

Satellite overhead imagery differs from many standard image classification challenges [18-19]. Overhead imagery offers no obvious up-and-down so augmentation [32] with rotation can offer diverse testing images that otherwise would not make sense for other datasets like fashion, digits, or alphabetic options. Similarly, the diversity of resolutions [21] offered from combined datasets and different satellites represent a noise and resolution variation that more standardized benchmarks can avoid in their domains. For leveraging previous MNIST work the major challenge in building a representative satellite dataset follows from a shortage of labelled data.

The largest crowd-sourced implementation (xView) labelled a million objects with bounding boxes [20] but offered an unbalanced count that mirrors the expectations of urban landscapes: buildings outnumber vehicles but together those two classes dominate the remaining 60+ classes. Creating a balanced dataset across just 10 object classes here involved fusing multiple sources, rescaling, cropping, or chipping to the desired object centred in a grayscale thumbnail. Another design consideration stems from the lack of realism to growing Overhead-MNIST with more non-trivial examples. For instance, using quarter rotations of each image and then brightening them by 10% would generate the comparable 70,000+ examples in MNIST. However, the entire globe might never offer that many stadiums (4,827 currently).

### 3.3. Mislabelling and Very Hard Samples

For large, multi-class benchmarks, previous work [33-34] has isolated mis-labelled or very hard samples. Crowd-sourcing the labelling (via Amazon Mechanical Turk or other platforms) highlights the differences in human opinions or alternately the diversity of data collection needed to establish ground-truth. This challenge appears even in carefully curated datasets like Fashion-MNIST, where real physical poses and photographers define the object classes and ontology from the Zalando catalogue. But given the top 5 algorithms in a state-of-the-art (SOTA) list differ by a fraction of a percent [35], we explore the very hard samples shown in Figure 7. For 1000 random samples, the labelling by humans identified 7 examples of tough samples (0.7%), either because the images contained a human model, were not front facing, were outside the 10 categories or combined two categories (shirt and trousers). The point of this exercise is less to nit-pick a tough labelling challenge but to note that some of this ontology noise is expected but may bear on the fine-tuning of certain algorithmic comparisons that follow, particularly for these highly popular ones [17] like MS-COCO, ImageNet, CIFAR-10 [22] and MNIST [1] families.

## 4. DISCUSSION AND CONCLUSIONS

The Overhead-MNIST dataset covers satellite imagery of 10 frequently seen object classes. The format corresponds closely to traditional digit recognition benchmarks, including images (compressed JPEG), CSV and binary (idx-ubyte). With pre-trained convolutional neural networks, transfer learning achieves 96.4% accuracy for multi-class identification. A more complete algorithm survey will follow as a companion to the benchmark dataset and preparation described here. With a future goal for better on-board processing, companion algorithmic studies will leverage the vast MNIST literature to shrink classifier models so they run on newer and smaller hardware.

## ACKNOWLEDGEMENTS

The author would like to thank the PeopleTec Technical Fellows program for encouragement and project assistance.